\documentclass[runningheads]{llncs}
\usepackage{graphicx}

\usepackage{tikz}
\usepackage{comment}
\usepackage{amsmath,amssymb} 
\usepackage{color}

\usepackage{latexsym}
\usepackage{epsfig}
\usepackage{graphicx}
\usepackage{amsmath,bm}
\usepackage{amssymb}
\usepackage{multirow}
\usepackage{verbatim}
\usepackage{enumitem}
\usepackage{booktabs,amsfonts,dcolumn}
\usepackage{hyperref}
\definecolor{deeppink}{rgb}{1.0, 0.08, 0.58}
\hypersetup{
    colorlinks=true,
    urlcolor=deeppink,
}
\usepackage{cleveref}
\makeatletter
\newcommand{\@chapapp}{\relax}%
\makeatother
\usepackage[title]{appendix}

\begin{document}
\pagestyle{headings}
\mainmatter
\def\ECCVSubNumber{2115}  

\title{Comprehensive Image Captioning \\via Scene Graph Decomposition} 

\titlerunning{Sub-graph Captioning}
\author{Yiwu Zhong\inst{1}\thanks{Work partially done while Yiwu Zhong was an intern at Tencent AI Lab, Bellevue.} 
\and
Liwei Wang\inst{2} 
\and
Jianshu Chen\inst{2} 
\and
Dong Yu\inst{2}
\and
Yin Li\inst{1}}
\authorrunning{Y. Zhong et al.}
\institute{University of Wisconsin-Madison, United States \and 
Tencent AI Lab, Bellevue, United States \\
\email{\{yzhong52, yin.li\}@wisc.edu},
\email{\{liweiwang, jianshuchen, dyu\}@tencent.com}}

\maketitle

\begin{abstract}
We address the challenging problem of image captioning by revisiting the representation of image scene graph. At the core of our method lies the decomposition of a scene graph into a set of sub-graphs, with each sub-graph capturing a semantic component of the input image. We design a deep model to select important sub-graphs, and to decode each selected sub-graph into a single target sentence. By using sub-graphs, our model is able to attend to different components of the image. Our method thus accounts for accurate, diverse, grounded and controllable captioning at the same time. We present extensive experiments to demonstrate the benefits of our comprehensive captioning model. Our method establishes new state-of-the-art results in caption diversity, grounding, and controllability, and compares favourably to latest methods in caption quality. Our project website can be found at \url{http://pages.cs.wisc.edu/~yiwuzhong/Sub-GC.html}.
\end{abstract}
\section{Introduction}

\begin{figure}[t]
	\centering
	\includegraphics[width=0.75\linewidth]{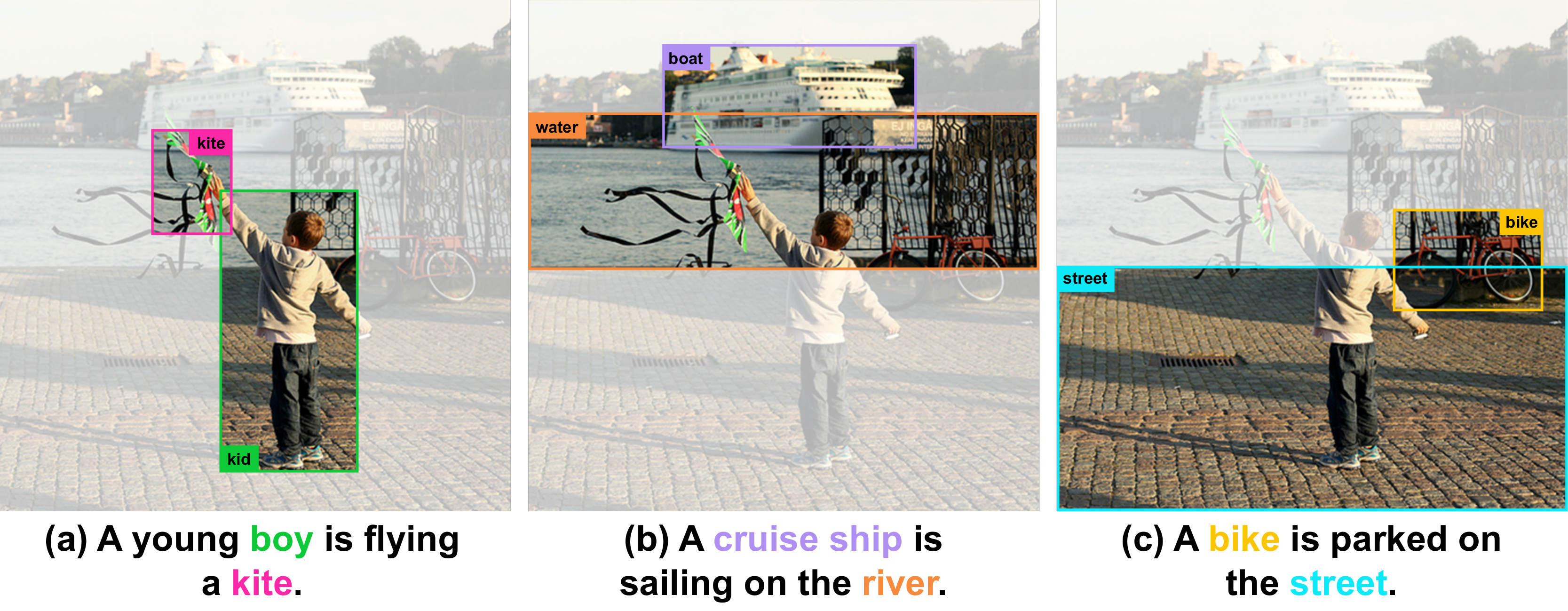}
    \caption{An example image with multiple scene components with each described by a distinct caption. {\it How can we design a model that can learn to identify and describe different components of an input image?}}
    \label{fig:teaser}
\end{figure}

It is an old saying that ``A picture is worth a thousand words''. Complex and sometimes multiple ideas can be conveyed by a single image. Consider the example in Fig.\ \ref{fig:teaser}. The image can be described by ``A boy is flying a kite'' when pointing to the boy and the kite, or depicted as ``A ship is sailing on the river'' when attending to the boat and the river. Instead, when presented with regions of the bike and the street, the description can be ``A bike parked on the street''. Humans demonstrate remarkable ability to summarize multiple ideas associated with different scene components of the same image. More interestingly, we can easily explain our descriptions by linking sentence tokens back to image regions. 

Despite recent progress in image captioning, most of current approaches are optimized for caption quality. These methods tend to produce generic sentences that are minorly reworded from those in the training set, and to ``look'' at regions that are irrelevant to the output sentence~\cite{das2017human,rohrbach-etal-2018-object}. Several recent efforts seek to address these issues, leading to models designed for individual tasks including diverse~\cite{wang2017diverse,deshpande2019fast}, grounded~\cite{selvaraju2019taking,zhou2019grounded} and controllable captioning~\cite{lu2018neural,cornia2019show}. However, no previous method exists that can address diversity, grounding and controllability at the same time---an ability seemingly effortless for we humans. 

We believe the key to bridge the gap is a semantic representation that can better link image regions to sentence descriptions. To this end, we propose to revisit the idea of image captioning using scene graph---a knowledge graph that encodes objects and their relationships. Our core idea is that such a graph can be decomposed into a set of sub-graphs, with each sub-graph as a candidate scene component that might be described by a unique sentence. Our goal is thus to design a model that can identify meaningful sub-graphs and decode their corresponding descriptions. A major advantage of this design is that diversity and controllability are naturally enabled by selecting multiple distinct sub-graphs to decode and by specifying a set of sub-graphs for sentence generation. 

Specifically, our method takes a scene graph extracted from an image as input. This graph consists of nodes as objects (nouns) and edges as the relations between pairs of objects (predicates). Each node or edge comes with its text and visual features. Our method first constructs a set of overlapping sub-graphs from the full graph. We develop a graph neural network that learns to select meaningful sub-graphs best described by one of the human annotated sentences. Each of the selected sub-graphs is further decoded into its corresponding sentence. This decoding process incorporates an attention mechanism on the sub-graph nodes when generating each token. Our model thus supports backtracking of generated sentence tokens into scene graph nodes and its image regions. Consequently, our method provides the \textit{first comprehensive model for generating accurate, diverse, and controllable captions that are grounded into image regions}. 

Our model is evaluated on MS-COCO Caption~\cite{chen2015microsoft} and Flickr30K Entities~\cite{plummer2015flickr30k} datasets. We benchmark the performance of our model on caption quality, diversity, grounding and controllability. Our results suggest that (1) top-ranked captions from our model achieve a good balance between quality and diversity, outperforming state-of-the-art methods designed for diverse captioning in both quality and diversity metrics and performing on par with latest methods optimized for caption quality in quality metrics; (2) our model is able to link the decoded tokens back into the image regions, thus demonstrating strong results for caption grounding; and (3) our model enables controllable captioning via the selection of sub-graphs, improving state-of-the-art results on controllability. We believe our work provides an important step for image captioning. 



\section{Related Work}

There has recently been substantial interest in image captioning. We briefly review relevant work on conventional image captioning, caption grounding, diverse and controllable captioning, and discuss related work on scene graph generation.

\noindent \textbf{Conventional Image Captioning}.
Major progress has been made in image captioning~\cite{hossain2019comprehensive}. An encoder-decoder model is often considered, where Convolutional Neural Networks (CNNs) are used to extract global image features, and Recurrent Neural Networks (RNNs) are used to decode the features into sentences~\cite{karpathy2015deep,vinyals2015show,donahue2015long,xu2015show,you2016image,lu2017knowing,rennie2017self,liu-etal-2018-simnet}. Object information has recently been shown important for captioning~\cite{yin2017obj2text,wang2018object}. Object features from an object detector can be combined with encoder-decoder models to generate high quality captions~\cite{Anderson2017up-down}.

Several recent works have explored objects and their relationships, encoded in the form of scene graphs, for image captioning~\cite{yao2018exploring,yang2019auto}. The most relevant work is~\cite{yao2018exploring}. Their GCN-LSTM model used a graph convolutional network (GCN)~\cite{kipf2016semi} to integrate semantic information in a scene graph. And a sentence is further decoded using features aggregated over the full scene graph. Similar to~\cite{yao2018exploring}, we also use a GCN for an input scene graph. However, our method learns to select sub-graphs within the scene graph, and to decode sentences from ranked sub-graphs instead of the full scene graph. This design allows our model to produce diverse and controllable sentences that are previously infeasible~\cite{yao2018exploring,yang2019auto,Anderson2017up-down}.

\noindent \textbf{Grounded Captioning}.
A major challenge of image captioning is that recent deep models might not focus on the same image regions as a human would when generating each word, leading to undesirable behaviors, e.g., object hallucination~\cite{rohrbach-etal-2018-object,das2017human}. Several recent work~\cite{xu2015show,Anderson2017up-down,selvaraju2019taking,zhou2019grounded,ma2019learning,johnson2016densecap,yang2017dense} has been developed to address the problem of grounded captioning---the generation of captions and the alignment between the generated words and image regions. Our method follows the weakly supervised setting for grounded captioning, where we assume that only the image-sentence pairs are known. Our key innovation is to use a sub-graph on an image scene graph for sentence generation, thus constraining the grounding within the sub-graph. 

Our work is also relevant to recent work on generating text descriptions of local image regions, also known as dense captioning~\cite{johnson2016densecap,yang2017dense,kim2019dense}. Both our work and dense captioning methods can create localized captions. The key difference is that our method aims to generate sentence descriptions of scene components that spans multiple image regions, while dense captioning methods focused on generating phrase descriptions for local regions~\cite{johnson2016densecap,yang2017dense} or pairs of local regions~\cite{kim2019dense}. 


\noindent \textbf{Diverse and Controllable Captioning}.
The generation of diverse and controllable image descriptions has also received considerable attention. Several approaches have been proposed for diverse captioning~\cite{shetty2017speaking,li2018generating,dai2017towards,vijayakumar2018diverse,wang2017diverse,deshpande2019fast,aneja2019sequential}. Wang et al.\ \cite{wang2017diverse} proposed a variational auto-encoder that can decode multiple diverse sentences from samples drawn from a latent space of image features. This idea was further extended by~\cite{aneja2019sequential}, where every word has its own latent space. Moreover, Deshpande et al.\ \cite{deshpande2019fast} proposed to generate various sentences controlled by part-of-speech tags. There is a few recent work on controllable captioning. Lu et al.\ \cite{lu2018neural} proposed to fill a generated sentence template with the concepts from an object detector. Cornia et al.\ \cite{cornia2019show} selected object regions using grounding annotations and then predicted textual chunks to generate diverse and controllable sentences. Similar to~\cite{cornia2019show}, we address diversity and controllability within the same model. Different from~\cite{cornia2019show}, our model is trained using only image-sentence pairs and can provide additional capacity of caption grounding.

\noindent \textbf{Scene Graph Generation}.
Image scene graph generation has received considerable attention, partially driven by large-scale scene graph datasets~\cite{krishna2017visual}. Most existing methods~\cite{lu2016visual,zhang2017visual,xu2017scenegraph,dai2017detecting,li2017scene,yang2018graph,li2017vip,zellers2018scenegraphs} start from candidate object regions given by an object detector and seek to infer the object categories and their relationships. By using a previous approach~\cite{zellers2018scenegraphs} to extract image scene graphs, we explore the decomposition of scene graphs into sub-graphs for generating accurate, diverse, and controllable captions. Similar graph partitioning problems have been previously considered in vision for image segmentation~\cite{shi2000normalized,felzenszwalb2004efficient} and visual tracking~\cite{tang2015subgraph,song2019end}, but has not been explored for image captioning.

\section{Method}

Given an input image $I$, we assume an image scene graph $G=(V,E)$ can be extracted from $I$, where $V$ represents the set of nodes corresponding to the detected objects (nouns) in $I$, and $E$ represents the set of edges corresponding to the relationships between pairs of objects (predicates). Our goal is to generate a set of sentences $C=\{C_j\}$ to describe different components of $I$ using the scene graph $G$. To this end, we propose to make use of the sub-graphs $\{G^s_i=(V^s_i, E^s_i)\}$ from $G$, where $V^s_i \subseteq V$ and $E^s_i \subseteq E$. Our method seeks to model the joint probability $P(S_{ij} = (G, G^s_i, C_j)|I)$, where $P(S_{ij}|I) = 1$ indicates that the sub-graph $G^s_i$ can be used to decode the sentence $C_j$. Otherwise, $P(S_{ij}|I) = 0$. We further assume that $P(S_{ij} |I)$ can be decomposed into three parts, given by
\begin{equation}\small
P(S_{ij}|I) = P(G|I) P(G^s_i | G, I) P(C_j | G^s_i, G, I).
\end{equation}

\begin{figure*}[t]
	\centering
	\includegraphics[width=0.85\linewidth]{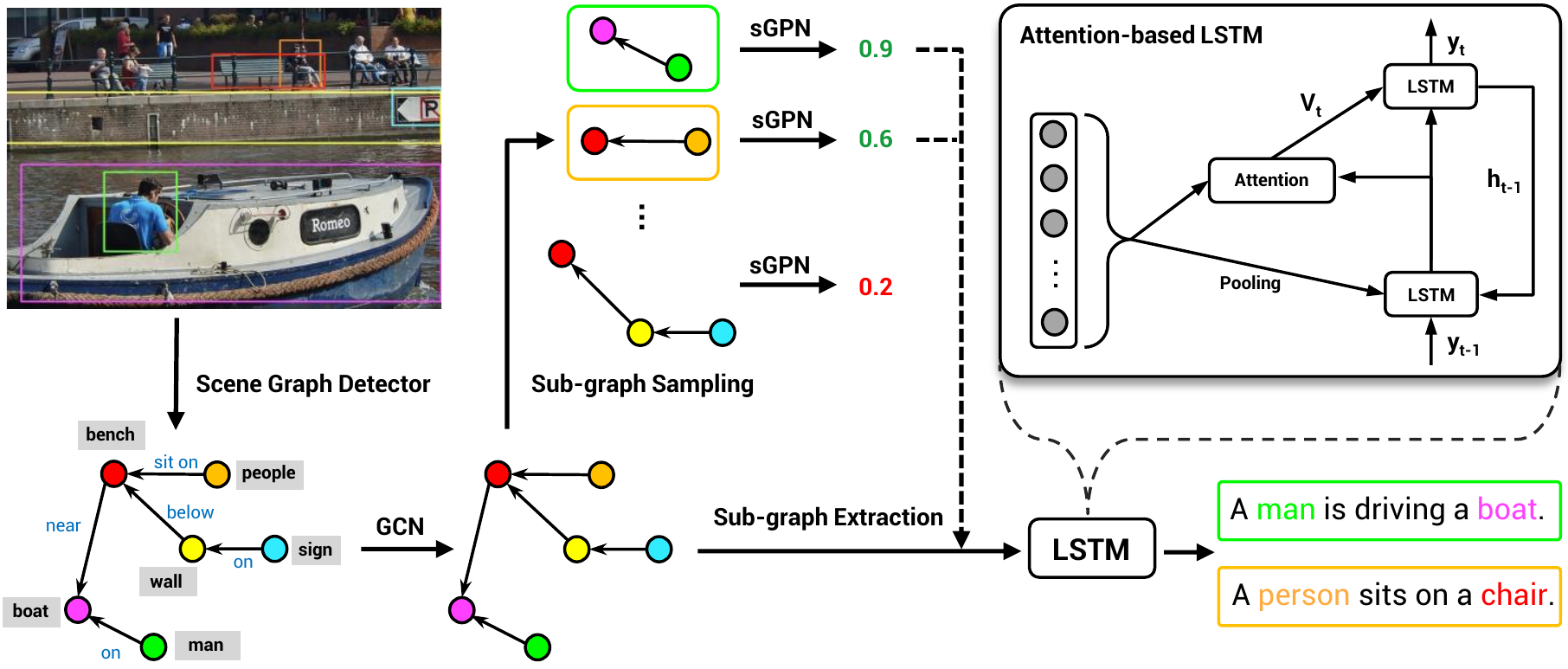}
    \caption{Overview of our method. Our method takes a scene graph extracted from an input image, and decomposes the graph into a set of sub-graphs. We design a sub-graph proposal network (sGPN) that learns to identify meaningful sub-graphs, which are further decoded by an attention-based LSTM for generating sentences and grounding sentence tokens into sub-graph nodes (image regions). By leveraging sub-graphs, our model enables accurate, diverse, grounded and controllable image captioning.}
    \label{fig:model_overview}
\end{figure*}

Intuitively, $P(G|I)$ extracts scene graph $G$ from an input image $I$. $P(G^s_i | G, I)$ decomposes the full graph $G$ into a diverse set of sub-graphs $\{G^s_i\}$ and selects important sub-graphs for sentence generation. Finally, $P(C_j | G^s_i, G, I)$ decodes a selected sub-graph $G^s_i$ into its corresponding sentence $C_j$, and also associates the tokens in $C_j$ to the nodes $V^s_i$ of the sub-graph $G^s_i$ (the image regions in $I$). Fig.\ \ref{fig:model_overview} illustrates our method. We now present details of our model.

\subsection{Scene Graph Detection and Decomposition}
Our method first extracts scene graph $G$ from image $I$ ($P(G|I)$) using MotifNet \cite{zellers2018scenegraphs}. MotifNet builds LSTMs on top of object detector outputs~\cite{ren2015faster} and produces a scene graph $G=(V,E)$ with nodes $V$ for common objects (nouns) and edges $E$ for relationship between pairs of objects (predicates), such as ``holding'',``behind'' or ``made of''. Note that $G$ is a directed graph, i.e., an edge must start from a subject noun or end at an object noun. Therefore, the graph $G$ is defined by a collection of subject-predicate-object triplets, e.g., kid playing ball. 

We further samples sub-graphs $\{G^s_i\}$ from the scene graph $G$ by using neighbor sampling~\cite{klusowski2018counting}. Specifically, we randomly select a set of seed nodes $\{S_i\}$ on the graph. The immediate neighbors of the seed nodes with the edges in-between define a sampled sub-graph. Formally, the sets of sub-graph nodes and edges are $V^s_i = S_i \cup \{N(v) | v \in S_i\}$ and $E^s_i = \{(v,u) | v \in S_i, u \in N(v)\}$ respectively, where $N(v)$ denotes the immediate neighbors of node $v$. Identical sub-graphs are removed to obtain the final set of sub-graphs $\{G^s_i=(V^s_i, E^s_i)\}$, which covers potential scene components in the input image $I$. 

\subsection{Sub-graph Proposal Network}
\label{Sub-graph Proposal Network}
Our next step is to identify meaningful sub-graphs that are likely to capture major scene components in the image ($P(G^s_i | G, I)$). Specifically, our model first combines visual and text features on the scene graph $G$, followed by an integration of contextual information within $G$ using a graph convolutional network, and finally a score function learned to rank sub-graphs $G^s_i$. 

\noindent \textbf{Scene Graph Representation}. Given a directed scene graph $G=(V,E)$, we augment its nodes and edges with visual and text features. For a node $v \in V$, we use both its visual feature extracted from image regions and the word embedding of its noun label. We denote the visual features as $\bm{x}_v^v \in \mathbb{R}^{d_v}$ and text features as $\bm{x}_v^e \in \mathbb{R}^{d_e}$. For an edge $e \in E$, we only use the embedding of its predicate label denoted as $\bm {x}_{e}^e \in \mathbb{R}^{d_e}$. Subscripts are used to distinguish node ($v$) and edge ($e$) features and superscripts to denote the feature type, i.e., visual features or text embedding. Visual and text features are further fused by projecting them into a common sub-space. This is done separately for node and edge features by
\begin{equation}\small
\bm x_v^f  =  {\rm ReLU}(\bm {W}_{f}^1 \bm {x}_{v}^v + \bm {W}_{f}^2 \bm {x}_{v}^{e}), \quad \bm x_e^f  =  \bm {W}_{f}^3 \bm {x}_{e}^e,
\end{equation}
where $\bm {W}_{f}^1 \in \mathbb{R}^{d_f \times d_{v}}$, $\bm {W}_{f}^2 \in \mathbb{R}^{d_f \times d_{e}}$ and $\bm {W}_{f}^3 \in \mathbb{R}^{d_f \times d_{e}}$ are learned projections.

\noindent \textbf{Graph Convolutional Network (GCN)}. After feature fusion and projection, we further model the context between objects and their relationships using a GCN. The GCN aggregates information from the neighborhood within the graph and updates node and edge features. With an proper ordering of the nodes and edges, we denote the feature matrix for nodes and edges as $\bm X_{v}^f = [\bm {x}_{v}^{f}] \in \mathbb{R}^{{d_f} \times |V|}$ and $\bm X_{e}^f = [\bm {x}_{e}^{f}] \in \mathbb{R}^{{d_f} \times |E|}$, respectively. The update rule of a single graph convolution is thus given by
\begin{equation}\small
\begin{split}
\bm {\hat X}_{v}^{f} &= \bm X_{v}^{f} + {\rm ReLU}(\bm W_{ps} \bm X_{e}^{f} \bm A_{ps}) + {\rm ReLU}(\bm W_{po} \bm X_{e}^{f} \bm A_{po}),\\
\bm {\hat X}_{e}^{f} &= \bm X_{e}^{f} + {\rm ReLU}(\bm W_{sp} \bm X_{v}^f \bm A_{sp}) + {\rm ReLU}(\bm W_{op} \bm X_{v}^f \bm A_{op}),
\end{split}
\end{equation}
where $\bm {W}_{ps}, \bm {W}_{po}, \bm {W}_{sp}, \bm {W}_{op} \in \mathbb{R}^{{d_f} \times {d_f}}$ are learnable parameters that link subject or object features (nouns) with predicate features. For example, $\bm {W}_{ps}$ connects between predicate features and subject features. $\bm A_{ps}, \bm A_{po} \in \mathbb{R}^{|E| \times |V|}$ are the normalized adjacency matrix (defined by $G$) between predicates and subjects, and between predicates and objects, respectively. For instance, a non-zero element in $\bm A_{ps}$ suggests a link between a predicate and a subject on the scene graph $G$. Similarly, $\bm A_{sp}, \bm A_{op} \in \mathbb{R}^{|V| \times |E|}$ are the normalized adjacency matrix between subjects and predicates, and between objects and predicates. $\bm A_{ps} \neq \bm A_{sp}^T$ due to the normalization of adjacency matrix.

Our GCN stacks several graph convolutions, and produces an output scene graph with updated node and edge features. We only keep the final node features ($\bm X^{u}_v = [\bm {x}^{u}_{v}], v \in V$) for subsequent sub-graph ranking, as the predicate information has been integrated using GCN.

\noindent \textbf{Sub-graph Score Function}. 
With the updated scene graph and the set of sampled sub-graphs, our model learns a score function to select meaningful sub-graphs for generating sentence descriptions. For each sub-graph, we index its node features as $\bm X_{i}^{s} = [\bm {x}^{u}_{v}], v \in V_i^s$ and construct a score function
\begin{equation} \label{eq:score} \small
s_{i} = \sigma (f(\Phi (\bm X_{i}^{s}))),
\end{equation} 
where $\Phi(\cdot)$ is a sub-graph readout function~\cite{xu2018powerful} that concatenates the max-pooled and mean-pooled node features on the sub-graph. $f(\cdot)$ is a score function realized by a two-layer multilayer perceptron (MLP). And $\sigma(\cdot)$ is a sigmoid function that normalizes the output score into the range of $[0, 1]$.

\noindent \textbf{Learning the Score Function}. The key challenge of learning the score function $f$ is the training labels. Our goal is to rank the sampled sub-graphs and select the best ones to generate captions. Thus, we propose to use ground-truth captions provided by human annotators to guide the learning. A sub-graph with most of the nodes matched to one of the ground-truth sentences should be selected. To this end, we recast the learning of the score function $f$ as training a binary classifier to distinguish between ``good'' (positive) and ``bad'' (negative) sub-graphs. Importantly, we design a matching score between a ground-truth sentence and a sampled sub-graph to generate the target binary labels. 

Specifically, given a sentence $C_j$ and a scene graph $G$, we extract a reference sub-graph on $G$ by finding the nodes on the graph $G$ that also appears in the sentence $C_j$ and including their immediate neighbor nodes. This is done by extracting nouns from the sentence $C_j$ using a part-of-speech tag parser~\cite{loper2002nltk}, and matching the nouns to the nodes on $G$ using LCH score~\cite{leacock1998using} derived from WordNet~\cite{miller1995wordnet}. This matching process is given by $\mathcal{M}(C_j, G)$. We further compute the node Intersection over Union (IoU) score between the reference sub-graph $\mathcal{M}(C_j, G)$ and each of the sampled sub-graph $G_I^s$ by 
\begin{equation}\small
IoU(G^s_i, C_j) = \frac {|G^s_i \cap \mathcal{M}(C_j, G)|} {|G^s_i \cup \mathcal{M}(C_j, G)|},
\label{IoU}
\end{equation}
where $\cap$ and $\cup$ are the intersection and union operation over sets of sub-graph nodes, respectively. The node IoU provides a matching score between the reference sentence $C_j$ and the sub-graph $G^s_i$ and is used to determine our training labels. We only consider a sub-graph as positive for training if its IoU with any of the target sentences is higher than a pre-defined threshold (0.75).

\noindent \textbf{Training Strategy}. A major issue in training is that we have many negative sub-graphs and only a few positive ones. To address this issue, a mini-batch of sub-graphs is randomly sampled to train our sGPN, where positive to negative ratio is kept as $1$:$1$. If a ground-truth sentence does not match any positive sub-graph, we use the reference sub-graph from $\mathcal{M}(C_j, G)$ as its positive sub-graph.

\subsection{Decoding Sentences from Sub-graphs}
Our final step is to generate a target sentence using features from any selected single sub-graph ($P(C_j | G^s_i, G, I)$). We modify the attention-based LSTM~\cite{Anderson2017up-down} for sub-graph decoding, as shown in Fig.\ \ref{fig:model_overview} (top right). Specifically, the model couples an attention LSTM and a language LSTM. The attention LSTM assigns each sub-graph node an importance score, further used by the language LSTM to generate the tokens. Specifically, at each time step $t$, the attention LSTM is given by $\bm h_{t}^{A} = LSTM_{Att}([\bm h_{t-1}^{L}, \bm e_{t}, \bm{x}_{i}^s])$, where $\bm h_{t-1}^{L}$ is the hidden state of the language LSTM at time $t-1$. $e_{t}$ is the word embedding of the input token at time $t$ and $\bm{x}^{s}_i$ is the sub-graph feature. Instead of averaging all region features as~\cite{Anderson2017up-down,yao2018exploring}, our model uses the input sub-graph feature, given by $\bm{x}^{s}_i = g(\Phi (\bm X_{i}^{s}))$, 
where $g(\cdot)$ is a two-layer MLP, $\Phi(\cdot)$ is the same graph readout unit in Eq.\ \ref{eq:score}.

Based on hidden states $\bm h_{t}^{A}$ and the node features $\bm X_{i}^{s} = [\bm {x}^{u}_{v}]$ in the sub-graph, an attention weight $a_{v,t}$ at time $t$ for node $v$ is computed by $a_{v,t} = \bm {w}_{a}^{T} {\rm tanh} (\bm {W}_{v} \bm {x}^{u}_{v} + \bm W_{h} \bm h_{t}^{A})$ with learnable weights $\bm W_{v}$, $\bm W_{h}$ and $\bm {w}_{a}$. A softmax function is further used to normalize $\bm a_{t}$ into $\bm \alpha_{t}$ defined on all sub-graph nodes at time $t$. We use ${\bm {\alpha}}_{t}$ to backtrack image regions associated with a decoded token for caption grounding. Finally, the hidden state of the attention LSTM $\bm h_{t}^{A}$ and the attention re-weighted sub-graph feature $\bm{\mathcal{V}}_{t} = \sum_{v}\alpha_{v,t} \bm {x}_{v}^{u}$ are used as the input of the language LSTM---a standard LSTM that decodes the next word.

\subsection{Training and Inference}\label{Training and Inference}
We summarize the training and inference schemes of our model.\\
\noindent \textbf{Loss Functions}. Our sub-graph captioning model has three parts: $P(G|I)$, $P(G^s_i | G, I)$, $P(C_j | G^s_i, G, I)$, where the scene graph generation ($P(G|I)$) is trained independently on Visual Genome~\cite{krishna2017visual}. For training, we combine two loss functions for $P(G^s_i | G, I)$ and $P(C_j | G^s_i, G, I)$. Concretely, we use a binary cross-entropy loss for the sub-graph proposal network ($P(G^s_i | G, I)$), and a multi-way cross-entropy loss for the attention-based LSTM model to decode the sentences ($P(C_j | G^s_i, G, I)$). The coefficient between the two losses is set to 1.\\
\noindent \textbf{Inference}. During inference, our model extracts the scene graph, samples sub-graphs and evaluates their sGPN scores. \textit{Greedy Non-Maximal Suppression (NMS)} is further used to filter out sub-graphs that largely overlap with others, and to keep sub-graphs with high sGPN scores. The overlapping between two sub-graphs is defined by the IoU of their nodes. We find that using NMS during testing helps to remove redundant captions and to promote diversity. 

After NMS, top-ranked sub-graphs are decoded using an attention-based LSTM. As shown in~\cite{luo2020analysis}, an \textit{optional top-K sampling}~\cite{fan2018hierarchical,radford2019language} can be applied during the decoding to further improve caption diversity. We disable top-K sampling for our experiments unless otherwise noticed. The final output is thus a set of sentences with each from a single sub-graph. By choosing which sub-graphs to decode, our model can control caption contents. Finally, we use attention weights in the LSTM to ground decoded tokens to sub-graph nodes (image regions).

\section{Experiments}
We now describe our implementation details and presents our results. We start with an ablation study (\ref{Ablation Study}) for different model components. Further, we evaluate our model across several captioning tasks, including accurate and diverse captioning (\ref{Accurate and Diverse Captioning}), grounded captioning (\ref{Grounded Captioning}) and controllable captioning (\ref{Controllable Captioning}).

\noindent \textbf{Implementation Details}. 
We used Faster R-CNN~\cite{ren2015faster} with ResNet-101~\cite{he2016deep} from~\cite{Anderson2017up-down} as our object detector. Based on detection results, Motif-Net~\cite{zellers2018scenegraphs} was trained on Visual Genome~\cite{krishna2017visual} with 1600/20 object/predicate classes. For each image, we applied the detector and kept 36 objects and 64 triplets in scene graph. We sampled 1000 sub-graphs per image and removed duplicate ones, leading to an average of 255/274 sub-graphs per image for MS-COCO~\cite{chen2015microsoft}/Flickr30K~\cite{plummer2015flickr30k}. We used 2048D visual features for image regions and 300D GloVe~\cite {pennington2014glove} embeddings for node and edge labels. These features were projected into 1024D, followed by a GCN with depth of 2 for feature transform and an attention LSTM (similar to~\cite{Anderson2017up-down}) for sentence decoding. For training, we used Adam~\cite{kingma2014adam} with initial learning rate of 0.0005 and a mini-batch of 64 images and 256 sub-graphs. Beam search was used in decoding with beam size 2, unless otherwise noted. 

\subsection{Ablation Study}
\label{Ablation Study}
We first conduct an ablation study of our model components, including the ranking function, the NMS threshold and the optional top-K sampling. We now describe the experiment setup and report the ablation results.

\begin{table*}[t]
\centering
\caption{Ablation study on sub-graph/sentence ranking functions, the NMS thresholds and the top-K sampling during decoding. We report results for both accuracy (B4 and C) and diversity (Distinct Caption, 1/2-gram). Our model is trained on the train set of COCO caption and evaluated on the the validation set, following M-RNN split~\cite{mao2014deep}.}
\resizebox{1.0\textwidth}{!}{%
\begin{tabular}{c|c|c|c|ccccc}
Model           & \begin{tabular}[c]{@{}c@{}}Ranking \\ Function\end{tabular} & NMS  & \begin{tabular}[c]{@{}c@{}}Top-K \\ Sampling\end{tabular} & B4   & C     & \begin{tabular}[c]{@{}c@{}}Distinct \\ Caption\end{tabular} & \begin{tabular}[c]{@{}c@{}}1-gram \\ (Best 5)\end{tabular} & \begin{tabular}[c]{@{}c@{}}2-gram \\ (Best 5)\end{tabular} \\ \hline
Sub-GC-consensus & consensus                                                   & 0.75 & N/A                                                       & 33.0 & 107.6 & 59.3\%                                                        & 0.25   & 0.32   \\
Sub-GC-sGPN           & sGPN                                                        & 0.75 & N/A                                                       & 33.4 & 108.7 & 59.3\%                                                        & 0.28   & 0.37   \\
Sub-GC          & sGPN+consensus                                              & 0.75 & N/A                                                       & 34.3 & 112.9 & 59.3\%                                                        & 0.28   & 0.37   \\ \hline
Sub-GC-consensus & consensus                                                   & 0.55 & N/A                                                       & 32.5 & 105.6 & 70.5\%                                                        & 0.27   & 0.36   \\
Sub-GC-sGPN           & sGPN                                                        & 0.55 & N/A                                                       & 33.4 & 108.7 & 70.5\%                                                        & 0.32   & 0.42   \\
Sub-GC          & sGPN+consensus                                              & 0.55 & N/A                                                       & 34.1 & 112.3 & 70.5\%                                                        & 0.32   & 0.42   \\ \hline
Sub-GC-S               & sGPN+consensus                                              & 0.55 & T=0.6,K=3                                                 & 31.8 & 108.7 & 96.0\%                                                        & 0.39   & 0.57    \\
Sub-GC-S               & sGPN+consensus                                                        & 0.55 & T=0.6,K=5                                                 & 30.9 & 106.1  & 97.5\%                                                        & 0.41   & 0.60   \\
Sub-GC-S              & sGPN+consensus                                                        & 0.55 & T=1.0,K=3                                                 & 28.4 & 100.7  & 99.2\%                                                        & 0.43   & 0.64   \\ 
\end{tabular}}
\label{tb:ablation}
\end{table*}

\noindent \textbf{Experiment Setup}.  
We follow the evaluation protocol from ~\cite{vijayakumar2018diverse,wang2017diverse,deshpande2019fast,aneja2019sequential} and report both accuracy and diversity results using the M-RNN split~\cite{mao2014deep} of MS-COCO Caption dataset~\cite{chen2015microsoft}. Specifically, this split has 118,287/4,000/1,000 images for train/val/test set, with 5 human labeled captions per image. We train the model on the train set and report the results on the \textit{val} set. For accuracy, we report top-1 accuracy out of the top 20 output captions, using BLEU-4~\cite{papineni2002bleu} and CIDEr~\cite{vedantam2015cider}. For diversity, we evaluate the percentage of distinct captions from 20 sampled output captions, and report 1/2-gram diversity of the best 5 sampled captions using a ranking function. Beam search was disabled for this ablation study. Table~\ref{tb:ablation} presents our results and we now discuss our results. 

\noindent \textbf{Ranking function} is used to rank output captions. Our sGPN provides a socre for each sub-graph and thus each caption. Our sGPN can thus be re-purposed as a ranking function. We compare sGPN with consensus re-ranking~\cite{devlin2015exploring,mao2014deep} widely used in the literature~\cite{wang2017diverse,deshpande2019fast,aneja2019sequential}. Moreover, we also experiment with applying consensus on top-scored captions (e.g., top-4) from sGPN (sGPN+consensus). Our sGPN consistently outperforms consensus re-ranking for both accuracy and diversity (+1.1 CIDEr and +12\% 1-gram with NMS=0.75). Importantly, consensus re-ranking is computational expensive, while our sGPN incurs little computational cost. Further, combining our sGPN with consensus re-ranking (sGPN+consensus) improves top-1 accuracy (+4.2 CIDEr with NMS=0.75). sGPN+consensus produces the same diversity scores as sGPN, since only one ranking function (sGPN) is used in diversity evaluation.

\noindent \textbf{NMS threshold} is used during inference to eliminate similar sub-graphs (see Section~\ref{Training and Inference}). We evaluate two NMS thresholds (0.55 and 0.75). For all ranking functions, a lower threshold (0.55) increases diversity scores (+8\%/+14\% 1-gram for consensus/sGPN) and has comparable top-1 accuracy, expect for consensus re-ranking (-2.0 CIDEr). Note that for our sGPN, the top-1 accuracy remains the same as the top-ranked sub-graph stays unchanged.

\noindent \textbf{Top-K sampling} is optionally applied during caption decoding, where each token is randomly drawn from the top $K$ candidates based on the normalized logits produced by a softmax function with temperature $T$. A small $T$ favors the top candidate and a large $K$ produces more randomness. We evaluate different combinations of $K$ and $T$. Using top-K sampling decreases the top-1 accuracy yet significantly increases all diversity scores (-3.6 CIDEr yet +22\% in 1-gram with T=0.6, K=3). The same trend was also observed in~\cite{luo2020analysis}.

\noindent \textbf{Our final model} (Sub-GC) combines sGPN and consensus re-ranking for ranking captions. We set NMS threshold to 0.75 for experiments focusing on the accuracy of top-1 caption (Table~\ref{tb:Karpathy_split}, \ref{tb:grounding}, \ref{tb:controllability}) and 0.55 for experiments on diversity (Table~\ref{tb:MRNN_split}). Top-K sampling is only enabled for additional results on diversity. 

\subsection{Accurate and Diverse Image Captioning}
\label{Accurate and Diverse Captioning}
\noindent \textbf{Dataset and Metric}.
Moving forward, we evaluate our final model for accuracy and diversity on MS-COCO caption \textit{test} set using M-RNN split~\cite{mao2014deep}. Similar to our ablation study, we report top-1 accuracy and diversity scores by selecting from a pool of top 20/100 output sentences. Top-1 accuracy scores include BLEU~\cite{papineni2002bleu}, CIDEr~\cite{vedantam2015cider}, ROUGE-L~\cite{lin2004rouge}, METEOR~\cite{banerjee2005meteor} and SPICE~\cite{anderson2016spice}. And diversity scores include distinct caption, novel sentences, mutual overlap (mBLEU-4) and n-gram diversity. Beam search was disabled for a fair comparison.

\noindent \textbf{Baselines}. We consider several latest methods designed for diverse and accurate captioning as our baselines, including Div-BS~\cite{vijayakumar2018diverse}, AG-CVAE~\cite{wang2017diverse}, POS~\cite{deshpande2019fast}, POS+Joint~\cite{deshpande2019fast} and Seq-CVAE~\cite{aneja2019sequential}. We compare our results of Sub-GC to these baselines in Table~\ref{tb:MRNN_split}. In addition, we include the results of our model with top-K sampling (Sub-GC-S), as well as human performance for references of diversity. 

\begin{table}[t]
\caption{Diversity and top-1 accuracy results on COCO Caption dataset (M-RNN split~\cite{mao2014deep}). Best-5 refers to the top-5 sentences selected by a ranking function. Note that Sub-GC and Sub-GC-S have same top-1 accuracy in terms of sample-20 and sample-100, since we have a sGPN score per sub-graph and global sorting is applied over all sampled sub-graphs. Our models outperform previous methods on both top-1 accuracy and diversity for the majority of the metrics.
}\label{tb:MRNN_split}
\centering
\resizebox{1.0\textwidth}{!}{%
\begin{tabular}{c|c|ccccc|cccccccc}
\multirow{3}{*}{Method} 
& \multirow{3}{*}{\#}
& \multicolumn{5}{c|}{\begin{tabular}[c]{@{}c@{}}Diversity\\ \end{tabular}}
& \multicolumn{8}{c}{\begin{tabular}[c]{@{}c@{}}Top-1 Accuracy\\ \end{tabular}} \\ \cline{3-15}
& 
& \multirow{2}{*}{\begin{tabular}[c]{@{}c@{}}Distinct\\ Caption ($\uparrow$)\end{tabular}} 
& \multirow{2}{*}{\begin{tabular}[c]{@{}c@{}}\#novel\\ (Best 5) ($\uparrow$)\end{tabular}} 
& \multirow{2}{*}{\begin{tabular}[c]{@{}c@{}}mBLEU-4\\ (Best 5) ($\downarrow$)\end{tabular}} 
& \multirow{2}{*}{\begin{tabular}[c]{@{}c@{}}1-gram\\ (Best 5) ($\uparrow$)\end{tabular}} 
& \multirow{2}{*}{\begin{tabular}[c]{@{}c@{}}2-gram\\ (Best 5) ($\uparrow$)\end{tabular}}
& \multirow{2}{*}{\begin{tabular}[c]{@{}c@{}}B1\\ \end{tabular}} 
& \multirow{2}{*}{\begin{tabular}[c]{@{}c@{}}B2\\ \end{tabular}} 
& \multirow{2}{*}{\begin{tabular}[c]{@{}c@{}}B3\\ \end{tabular}} 
& \multirow{2}{*}{\begin{tabular}[c]{@{}c@{}}B4\\ \end{tabular}} 
& \multirow{2}{*}{\begin{tabular}[c]{@{}c@{}}C\\ \end{tabular}} 
& \multirow{2}{*}{\begin{tabular}[c]{@{}c@{}}R\\ \end{tabular}} 
& \multirow{2}{*}{\begin{tabular}[c]{@{}c@{}}M\\ \end{tabular}} 
& \multirow{2}{*}{\begin{tabular}[c]{@{}c@{}}S\\ \end{tabular}} \\
&  &    &      &      &      &     &  &  &    &      &      &      &   &  \\ \hline

Div-BS~\cite{vijayakumar2018diverse} & \multirow{5}{*}{20} & 100\% & 3106 & 81.3 & 0.20 & 0.26 & 72.9 & 56.2 & 42.4 & 32.0 & 103.2  & 53.6 & 25.5 & 18.4  \\
AG-CVAE~\cite{wang2017diverse}       & & 69.8\% & 3189 & 66.6 & 0.24 & 0.34 & 71.6 & 54.4 & 40.2 & 29.9 & 96.3  & 51.8 & 23.7 & 17.3  \\
POS~\cite{deshpande2019fast}         & & 96.3\% & 3394 & 63.9 & 0.24 & 0.35 & 74.4 & 57.0 & 41.9 & 30.6 & 101.4 & 53.1 & 25.2 & 18.8  \\
POS+Joint~\cite{deshpande2019fast}   & & 77.9\% & 3409 & 66.2 & 0.23 & 0.33 & 73.7 & 56.3 & 41.5 & 30.5 & 102.0 & 53.1 & 25.1 & 18.5  \\
Sub-GC          & & 71.1\%	&3679 &67.2 &0.31	&0.42 & \textbf{77.2} & \textbf{60.9} & \textbf{46.2} & \textbf{34.6} & \textbf{114.4} & \textbf{56.1} & \textbf{26.9} & \textbf{20.0}  \\  \hline
Seq-CVAE~\cite{aneja2019sequential}  & \multirow{2}{*}{20} & 94.0\% & \textbf{4266} & 52.0 & 0.25 & 0.54 & 73.1 & 55.4 & 40.2 & 28.9 & 100.0 & 52.1 & 24.5 & 17.5  \\
Sub-GC-S    & & \textbf{96.2\%}	&4153	&\textbf{36.4}	&\textbf{0.39}	&\textbf{0.57} & 75.2 & 57.6 & 42.7 & 31.4 & 107.3 & 54.1 & 26.1 & 19.3       \\
\hline \hline

Div-BS~\cite{vijayakumar2018diverse} & \multirow{5}{*}{100} & 100\% & 3421 & 82.4 & 0.20 & 0.25 & 73.4 & 56.9 & 43.0 & 32.5 & 103.4 & 53.8 & 25.5 & 18.7  \\
AG-CVAE~\cite{wang2017diverse}       & & 47.4\% & 3069 & 70.6 & 0.23   & 0.32 & 73.2 & 55.9 & 41.7 & 31.1 & 100.1 & 52.8 & 24.5 & 17.9  \\
POS~\cite{deshpande2019fast}         & & 91.5\% & 3446 & 67.3 & 0.23   & 0.33 & 73.7 & 56.7 & 42.1 & 31.1 & 103.6 & 53.0 & 25.3 & 18.8  \\
POS+Joint~\cite{deshpande2019fast}   & & 58.1\% & 3427 & 70.3 & 0.22   & 0.31 & 73.9 & 56.9 & 42.5 & 31.6 & 104.5 & 53.2 & 25.5 & 18.8  \\
Sub-GC           & & 65.8\%	&3647	&69.0	&0.31	&0.41 & \textbf{77.2} & \textbf{60.9} & \textbf{46.2} & \textbf{34.6} & \textbf{114.4} & \textbf{56.1} & \textbf{26.9} & \textbf{20.0}  \\ \hline
Seq-CVAE~\cite{aneja2019sequential}  &\multirow{2}{*}{100} & 84.2\% & \textbf{4215} & 64.0 & 0.33   & 0.48 & 74.3 & 56.8 & 41.9 & 30.8 & 104.1 & 53.1 & 24.8 & 17.8  \\
Sub-GC-S    & & \textbf{94.6\%}	&4128	&\textbf{37.3}	&\textbf{0.39}	&\textbf{0.57} & 75.2 & 57.6 & 42.7 & 31.4 & 107.3 & 54.1 & 26.1 & 19.3       \\
\hline \hline
Human     & 5 & 99.8\% & - & 51.0 & 0.34   & 0.48 & - & - & - & - & - & - & - & -  \\
\end{tabular} 
}
\newline
\caption{Comparison to accuracy optimized models on COCO caption dataset using Karpathy split~\cite{karpathy2015deep}. Our Sub-GC compares favorably to the latest methods.}
\label{tb:Karpathy_split}
\centering
\begin{tabular}[t]{c|ccccccc}
Method           & B1   & B4   & C     & R    & M    & S    \\ \hline
Up-Down~\cite{Anderson2017up-down}                 & 77.2 & 36.2 & 113.5 & 56.4 & 27.0 & 20.3 \\
GCN-LSTM~\cite{yao2018exploring}        & 77.3 & 36.8 & 116.3 & 57.0 & \textbf{27.9} & \textbf{20.9} \\
SGAE~\cite{yang2019auto}          & \textbf{77.6} & \textbf{36.9} & \textbf{116.7} & \textbf{57.2} & 27.7 & \textbf{20.9} \\ \hline
Full-GC          & 76.7 & \textbf{36.9} & 114.8 & 56.8 & \textbf{27.9} & 20.8 \\\
Sub-GC   & 76.8 & 36.2 & 115.3 & 56.6 & 27.7 & 20.7 \\ \hline
Sub-GC-oracle           & 90.7 & 59.3 & 166.7 & 71.5 & 40.1 & 30.1 \\
\end{tabular}
\end{table}

\noindent \textbf{Diversity Results}. For the majority of the diversity metrics, our model Sub-GC outperforms previous methods (+8\% for novel sentences and +29\%/20\% for 1/2-gram with 20 samples), except the most recent Seq-CVAE. Upon a close inspection of Seq-CVAE model, we hypothesis that Seq-CVAE benefits from sampling tokens at each time step. It is thus meaningful to compare our model using top-K sampling (Sub-GC-S) with Seq-CVAE. Sub-GC-S outperforms Seq-CVAE in most metrics (+18\%/19\% for 1/2-gram with 100 samples) and remains comparable for the metric of novel sentences (within 3\% difference).

\noindent \textbf{Accuracy Results}. We notice that the results of our sub-graph captioning models remain the same with increased number of samples. This is because our outputs follow a fixed rank from sGPN scores. Our Sub-GC outperforms all previous methods by a significant margin. Sub-GC achieves +2.6/2.1 in B4 and +11.2/9.9 in CIDEr when using 20/100 samples in comparison to previous best results. Moreover, while achieving best diversity scores, our model with top-K sampling (Sub-GC-S) also outperforms previous methods in most accuracy metrics (+0.8/0.9 in B1 and +4.1/2.8 in CIDEr when using 20/100 samples) despite its decreased accuracy from Sub-GC.

\noindent \textbf{Comparison to Accuracy Optimized Captioning models}. We conduct further experiments to compare the top ranked sentence from ou Sub-GC against the results of latest captioning models optimized for accuracy, including Up-Down~\cite{Anderson2017up-down}, GCN-LSTM~\cite{yao2018exploring} and SGAE~\cite{yang2019auto}. All these previous models can only generate a single sentence, while our method (Sub-GC) can generate a set of diverse captions. As a reference, we consider a variant of our model (Full-GC) that uses a full scene graph instead of sub-graphs to decode sentences. Moreover, we include an upper bound of our model (Sub-GC-oracle) by assuming that we have an oracle ranking function, i.e., always selecting the maximum scored sentence for each metric. All results are reported on Karpathy split~\cite{karpathy2015deep} of COCO dataset without using reinforcement learning for score optimization~\cite{rennie2017self}.

Our results are shown in Table \ref{tb:Karpathy_split}. Our Sub-GC achieves comparable results (within 1-2 points in B4/CIDEr) to latest methods (Up-Down, GCN-LSTM and SGAE). We find that the results of our sub-graph captioning model is slightly worse than those models using the full scene graph, e.g., Full-GC, GCN-LSTM and SGAE. We argue that this minor performance gap does not diminish our contribution, as our model offers new capacity for generating diverse, controllable and grounded captions. Notably, our best case (Sub-GC-oracle) outperforms all other methods for all metrics by a very large margin (+22.4 in B4 and +50.0 in CIDEr). These results suggest that at least one high-quality caption exists among the sentences decoded from the sub-graphs. It is thus possible to generate highly accurate captions if there is a way to select this ``good'' sub-graph.

\begin{table}[t]
\caption{Grounded captioning results on Flickr30K Entities~\cite{plummer2015flickr30k}. Our method (Sub-GC) outperforms previous weakly supervised methods.}\label{tb:grounding}
\centering
\begin{tabular}[t]{c|cc|ccccc}
\multirow{2}{*}{Method} & \multicolumn{2}{c|}{Grounding Evaluation} & \multicolumn{5}{c}{Caption Evaluation} \\ \cline{2-8} 
& F1 all   & F1 loc    & B1    & B4   & C     & M     & S     \\ \hline
GVD \cite{zhou2019grounded}            & 3.88                & 11.70               & 69.2   & 26.9  &60.1   &  22.1  & 16.1  \\
Up-Down \cite{Anderson2017up-down}                & 4.14                & 12.30               & 69.4   & 27.3  & 56.6  & 21.7 & 16.0  \\
Cyclical \cite{ma2019learning}               & 4.98                & 13.53               & 69.9   & 27.4  &  61.4  & 22.3 & 16.6  \\ \hline
Full-GC  & 4.90            & 13.08         & 69.8  & \textbf{29.1}  & \textbf{63.5}  & \textbf{22.7}  & \textbf{17.0}  \\  
Sub-GC              & \textbf{5.98}                & \textbf{16.53}               & \textbf{70.7}   & 28.5  &61.9   & 22.3  & 16.4  \\  \hline \hline
GVD (Sup.)~\cite{zhou2019grounded}             & 7.55                & 22.20               & 69.9   & 27.3  &62.3    & 22.5 & 16.5  \\
\end{tabular}%
\newline
\caption{Controllable captioning results on Flickr30K Entities~\cite{plummer2015flickr30k}. With weak supervision, our Sub-GC compares favorably to previous methods. With strong supervision, our Sub-GC (Sup.) achieves the best results.} \label{tb:controllability}
\centering
\begin{tabular}[t]{c|ccccccc}
Method & B1   & B4  &  C   & R    &M     & S    & IoU  \\ \hline
NBT~\cite{lu2018neural} (Sup.)    & -    & 8.6 & 53.8 & 31.9 &13.5  & 17.8 & 49.9  \\
SCT~\cite{cornia2019show} (Sup.)   & 33.1 & 9.9 & 67.3 & 35.3 & 14.9 & 22.2 & 52.7 \\ \hline
Sub-GC   & 33.6 & 9.3 & 57.8 & 32.5 &14.2  & 18.8 & 50.6  \\
Sub-GC (Sup.)  & \textbf{36.2} & \textbf{11.2} &  \textbf{73.7} & \textbf{35.5} & \textbf{15.9} & \textbf{22.2} & \textbf{54.1}  \\
\end{tabular}
\end{table}

\subsection{Grounded Image Captioning}
\label{Grounded Captioning}

Moreover, we evaluate our model for grounded captioning. We describe the dataset and metric, introduce our setup and baselines, and discuss our results. 

\noindent \textbf{Dataset and Metric}. We use Flickr30k Entities~\cite{plummer2015flickr30k} for grounded captioning. Flickr30k Entities has 31K images, with 5 captions for each image. The dataset also includes 275k annotated bounding boxes associated with the phrases in corresponding captions. We use the data split from~\cite{karpathy2015deep}. To evaluate the grounding performance, we follow the protocol in GVD~\cite{zhou2019grounded}. We report both $F1_{all}$ and $F1_{loc}$. $F1_{all}$ considers a region prediction as correct if the object word is correctly predicted and the box is correctly localized. On the other hand $F1_{loc}$ only accounts for localization quality. Moreover, we report the standard BLEU~\cite{papineni2002bleu}, CIDEr~\cite{vedantam2015cider}, METEOR~\cite{banerjee2005meteor} and SPICE~\cite{anderson2016spice} scores for caption quality. 

\noindent \textbf{Experiment Setup and Baselines}. For this experiment, we only evaluate the top-ranked sentence and its grounding from our model. We select the node on the sub-graph with maximum attention weight when decoding a noun word, and use its bounding box as the grounded region. Our results are compared to a strong set of baselines designed for weakly supervised grounded captioning, including weakly supervised GVD~\cite{zhou2019grounded}, Up-Down~\cite{Anderson2017up-down} and a concurrent work Cyclical~\cite{ma2019learning}. We also include reference results from fully supervised GVD~\cite{zhou2019grounded} that requires ground-truth matching pairs for training, and our Full-GC that decode a sentence from a full graph.

\begin{figure}[t]
	\centering
	\includegraphics[width=0.8\linewidth]{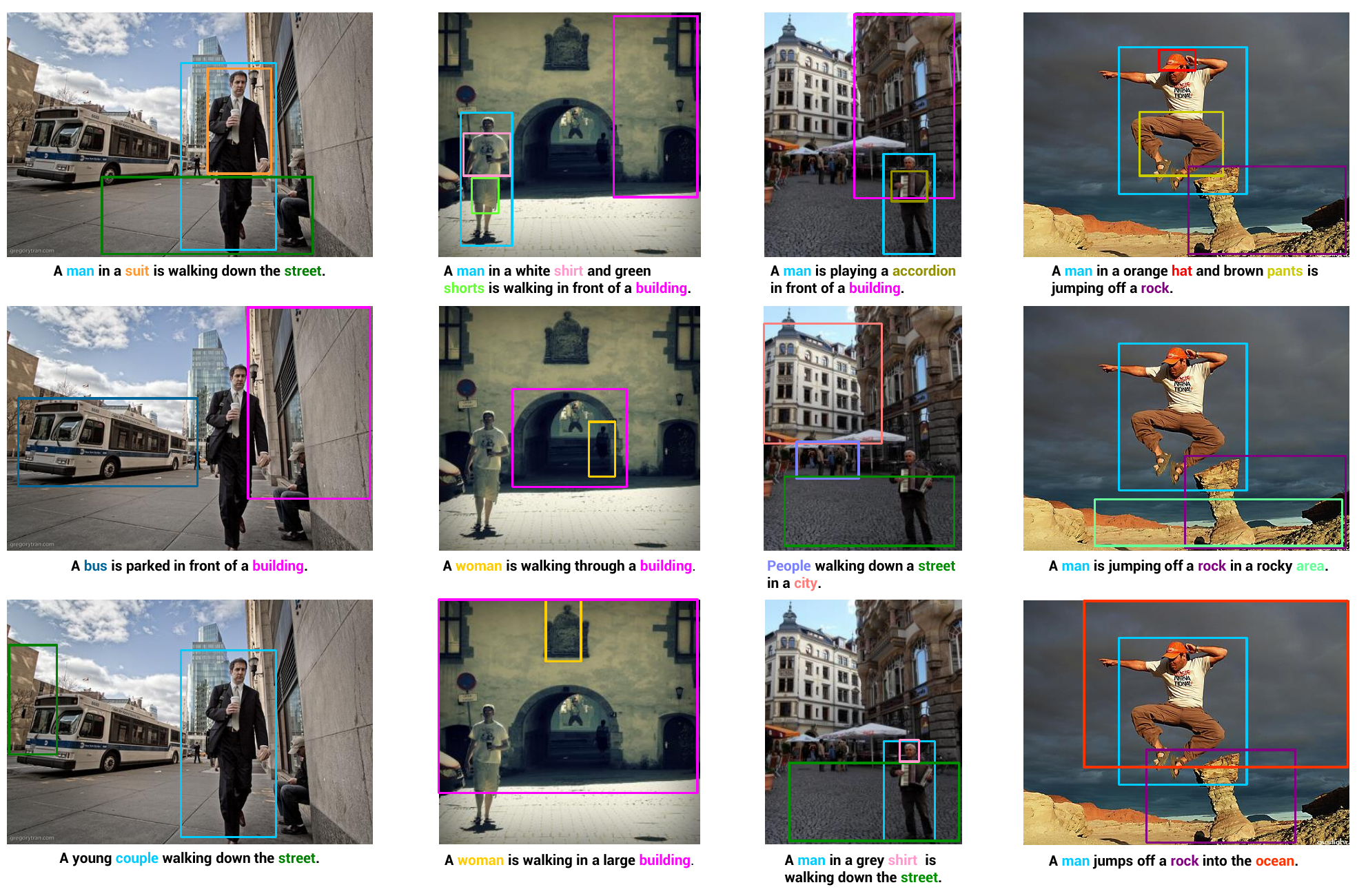}
    \caption{Sample results of our Sub-GC on Flickr30k Entities test set. Each column shows three captions with their region groundings decoded from different sub-graphs for an input image. The first two rows are successful cases and the last row is the failure case. These sentences can describe different parts of the images. Each generated noun and its grounding bounding box are highlighted in the same color.}
    \label{fig:grounding_visualization}
\end{figure}

\noindent \textbf{Results}.
Our results are presented in Table~\ref{tb:grounding}. Among all weakly supervised methods, our model achieves the best F1 scores for caption grounding. Specifically, our sub-graph captioning model (Sub-GC) outperforms previous best results by +1.0 for $F1_{all}$ and +3.0 for $F1_{loc}$, leading to a relative improvement of 20\% and 22\% for $F1_{all}$ and $F1_{loc}$, respectively. Our results also have the highest captioning quality (+1.1 in B4 and +0.5 in CIDEr). We conjecture that constraining the attention to the nodes of a sub-graph helps to improve the grounding. Fig.\ \ref{fig:grounding_visualization} shows sample results of grounded captions. Not surprisingly, the supervised GVD outperforms our Sub-GC. Supervised GVD can be considered as an upper bound for all other methods, as it uses grounding annotations for training. Comparing to our Full-GC, our Sub-GC is worse on captioning quality (-0.6 in B4 and -1.6 in CIDEr) yet has significant better performance for grounding (+1.1 in $F1_{all}$ and +3.5 in $F1_{loc}$).

\subsection{Controllable Image Captioning}
\label{Controllable Captioning}
Finally, we report results on controllable image captioning. Again, we describe our experiments and present the results. 

\noindent \textbf{Dataset and Metric}. Same as grounding, we consider Flickr30k Entities~\cite{plummer2015flickr30k} for controllable image captioning and use the data split~\cite{karpathy2015deep}. We follow evaluation protocol developed in~\cite{cornia2019show}. Specifically, the protocol assumes that an image and a set of regions are given as input, and evaluates a decoded sentence against one or more target ground-truth sentences. These ground-truth sentences are selected from captions by matching the sentences tokens to object regions in the image. Standard captioning metrics are considered (BLEU~\cite{papineni2002bleu}, CIDEr~\cite{vedantam2015cider}, ROUGE-L~\cite{lin2004rouge}, METEOR~\cite{banerjee2005meteor} and SPICE~\cite{anderson2016spice}), yet the ground-truth is different from conventional image captioning. Moreover, the IoU of the nouns between the predicted and the target sentence is also reported as~\cite{cornia2019show}.

\noindent \textbf{Experiment Setup and Baselines}. We consider (1) our Sub-GC trained with only image-sentence pairs; and (2) a supervised Sub-GC trained with ground-truth pairs of region sets and sentences as~\cite{cornia2019show}. Both models follow the same inference scheme, where input controlled set of regions are converted into best matching sub-graphs for sentence decoding. However, supervised Sub-GC uses these matching during training. We compare our results to recent methods developed for controllable captioning, including NBT~\cite{lu2018neural} and SCT~\cite{cornia2019show}. NBT and SCT are trained with matching pairs of region sets and sentences same as our supervised Sub-GC. Results are reported without using reinforcement learning.

\noindent \textbf{Results}. The results are shown in Table~\ref{tb:controllability}. Our models demonstrate strong controllability of the output sentences. Specifically, our supervised Sub-GC outperforms previous supervised methods (NBT and SCT) by a significant margin. Comparing to previous best SCT, our results are +1.3 in B4, +6.4 in CIDEr and +1.4 in IoU. Interestingly, our vanilla model has comparable performance to previous methods, even if it is trained with only image sentence pairs. These results provide further supports to our design of using sub-graphs for image captioning.

\section{Conclusion}
We proposed a novel image captioning model by exploring sub-graphs of image scene graph. Our key idea is to select important sub-graphs and only decode a single target sentence from a selected sub-graph. We demonstrated that our model can generate accurate, diverse, grounded and controllable captions. Our method thus offers the first comprehensive model for image captioning. Moreover, our results established new state-of-the-art in diverse captioning, grounded captioning and controllable captioning, and compared favourably to latest method for caption quality. We hope our work can provide insights into the design of explainable and controllable models for vision and language tasks.

\noindent \textbf{Acknowledgment}.
The work was partially developed during the first author's internship at Tencent AI Lab and further completed at UW-Madison. YZ and YL acknowledge the support by the UW VCRGE with funding from WARF.

%
%
\bibliographystyle{splncs04}
\bibliography{egbib}

\newpage
\begin{appendices}
\renewcommand{\thesection}{Appendix \Alph{section}}  

\section{Additional Implementation Details}

We present additional implementation details not covered in the main paper.

\noindent \textbf{Network Architecture}. $f(\cdot)$ in Eq.\ 4 took input features with a dimension of 2048 (D=2048), projected them into a vector (D=512), and outputted a scalar. Moreover, $g(\cdot)$ in Eq.\ 6 was a two-layer fully connected network that first projects input features (D=2048) to D=512 and then D=2048. All GCN layers transformed the input features (e.g., node and edge features with D=1024) to a feature dimension D=1024. The LSTMs used in our model followed the same architecture as~\cite{Anderson2017up-down}.

\noindent \textbf{Inference}. For consensus re-ranking, we used global image features from ResNet-101~\cite{he2016deep} pre-trained on ImageNet~\cite{deng2009imagenet}.



\section{More Qualitative Results}

\begin{figure*}[t]
	\centering
	\includegraphics[width=0.85\linewidth]{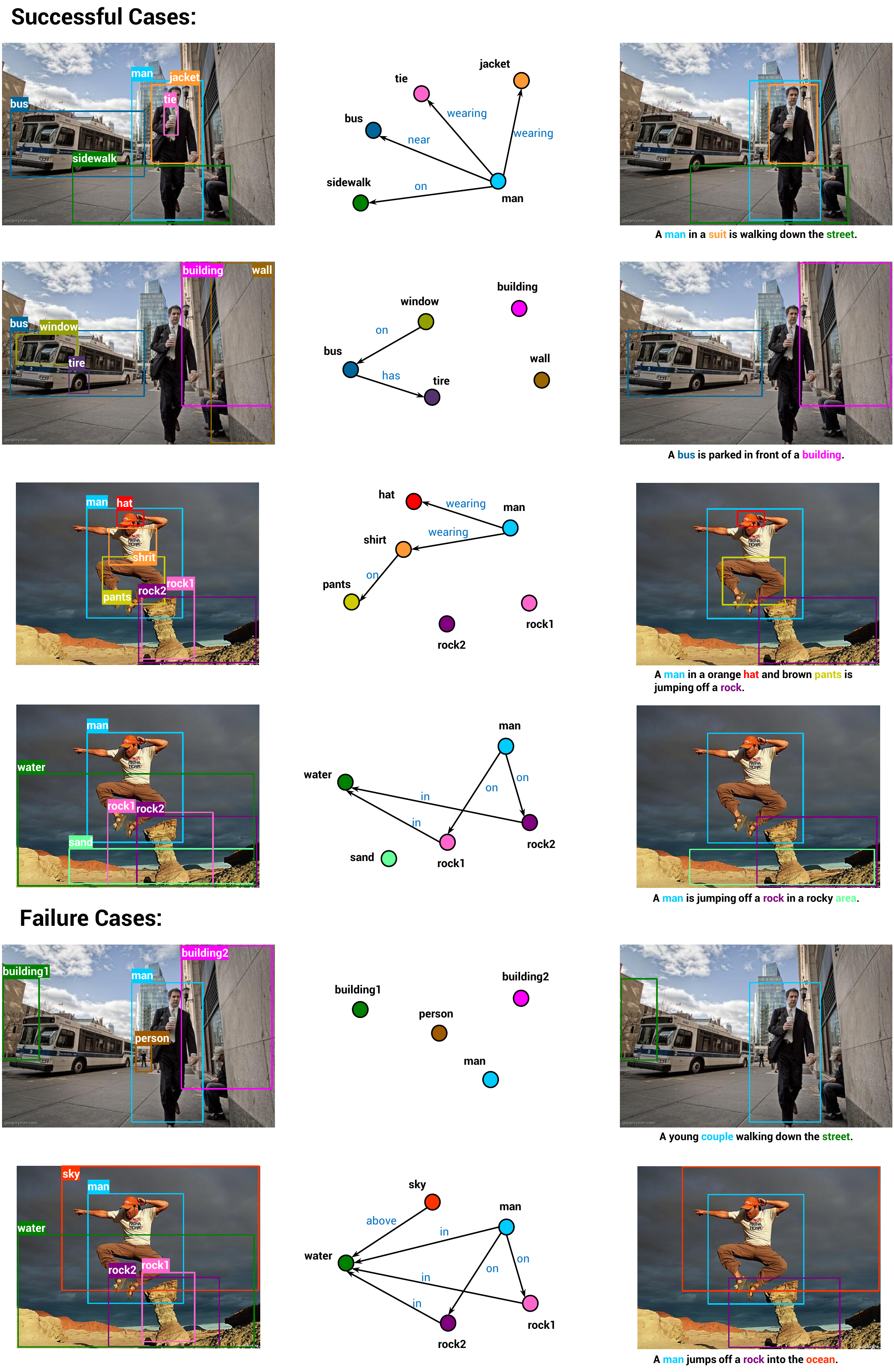}
    \caption{Diverse and grounded captioning results on Flickr30k Entities test set (Part 1). Each row presents the result from a single sub-graph. From left to right: input image and detected objects associated with the sub-graph, sub-graph nodes and edges used to decode the sentences, and the generated sentence grounded into image regions. Decoded nouns and their corresponding grounding regions are shown in the same color.}
    \label{fig:subgraph_grounding_visualization_part1}
\end{figure*}

\begin{figure*}[t]
	\centering
	\includegraphics[width=0.85\linewidth]{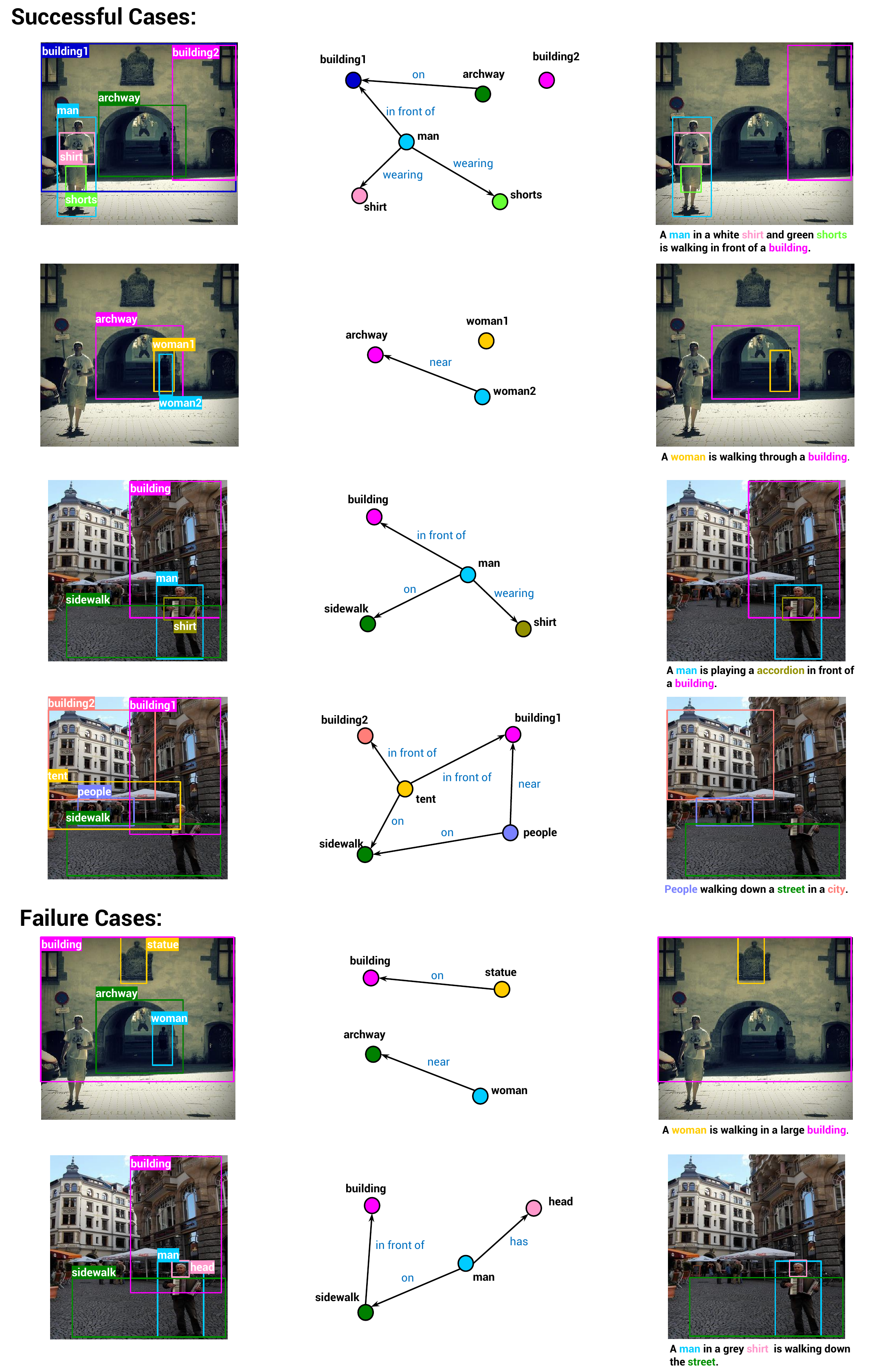}
    \caption{Diverse and grounded captioning results on Flickr30k Entities test set (Part 2). Each row presents the result from a single sub-graph. From left to right: input image and detected objects associated with the sub-graph, sub-graph nodes and edges used to decode the sentences, and the generated sentence grounded into image regions. Decoded nouns and their corresponding grounding regions are shown in the same color.}
    \label{fig:subgraph_grounding_visualization_part2}
\end{figure*}

\begin{figure*}[t]
	\centering
	\includegraphics[width=0.85\linewidth]{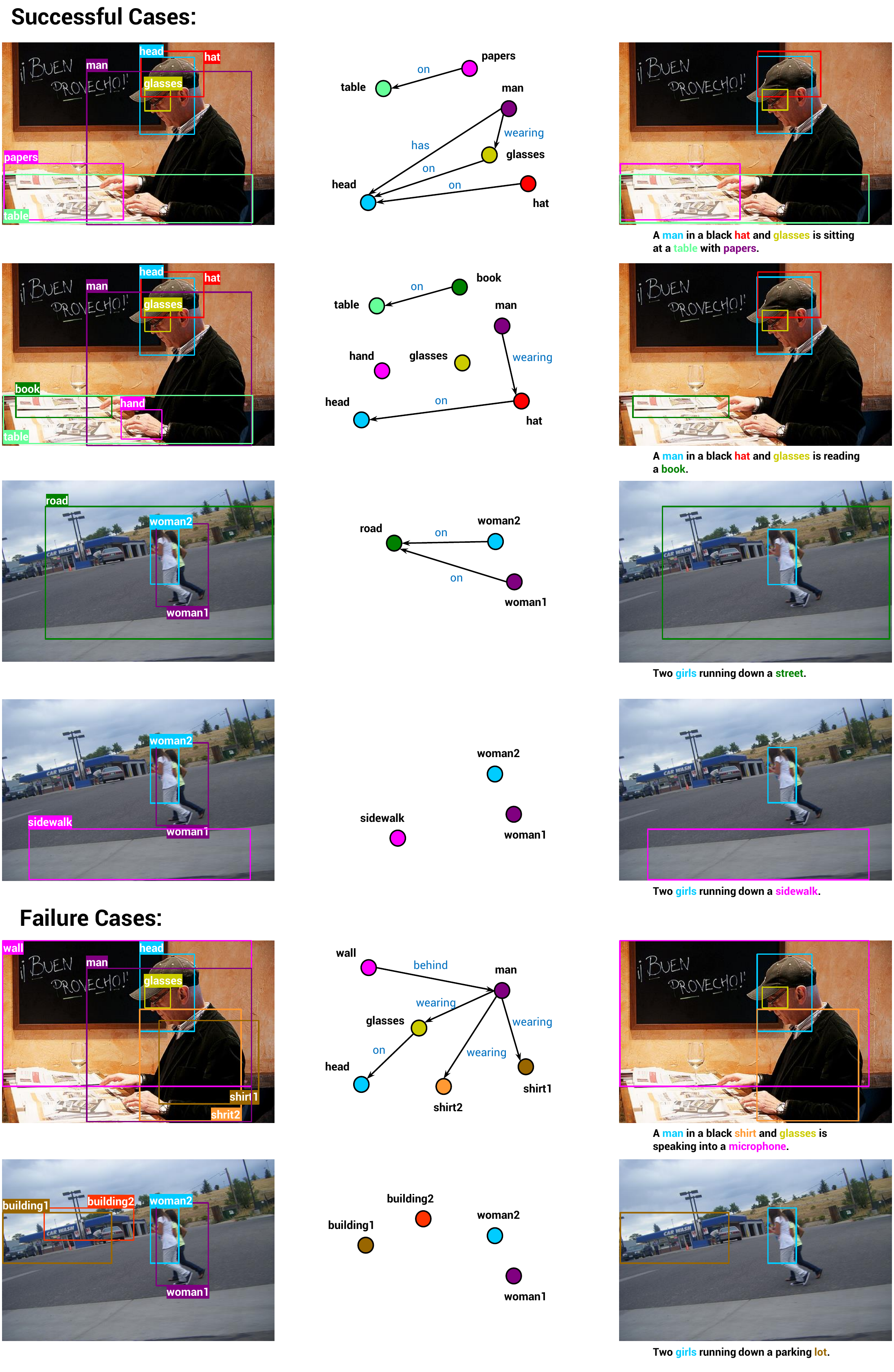}
    \caption{Diverse and grounded captioning results on Flickr30k Entities test set (Part 3). Each row presents the result from a single sub-graph. From left to right: input image and detected objects associated with the sub-graph, sub-graph nodes and edges used to decode the sentences, and the generated sentence grounded into image regions. Decoded nouns and their corresponding grounding regions are shown in the same color.}
    \label{fig:subgraph_grounding_visualization_part3}
\end{figure*}

\begin{figure*}[t]
	\centering 
	\includegraphics[width=0.85\linewidth]{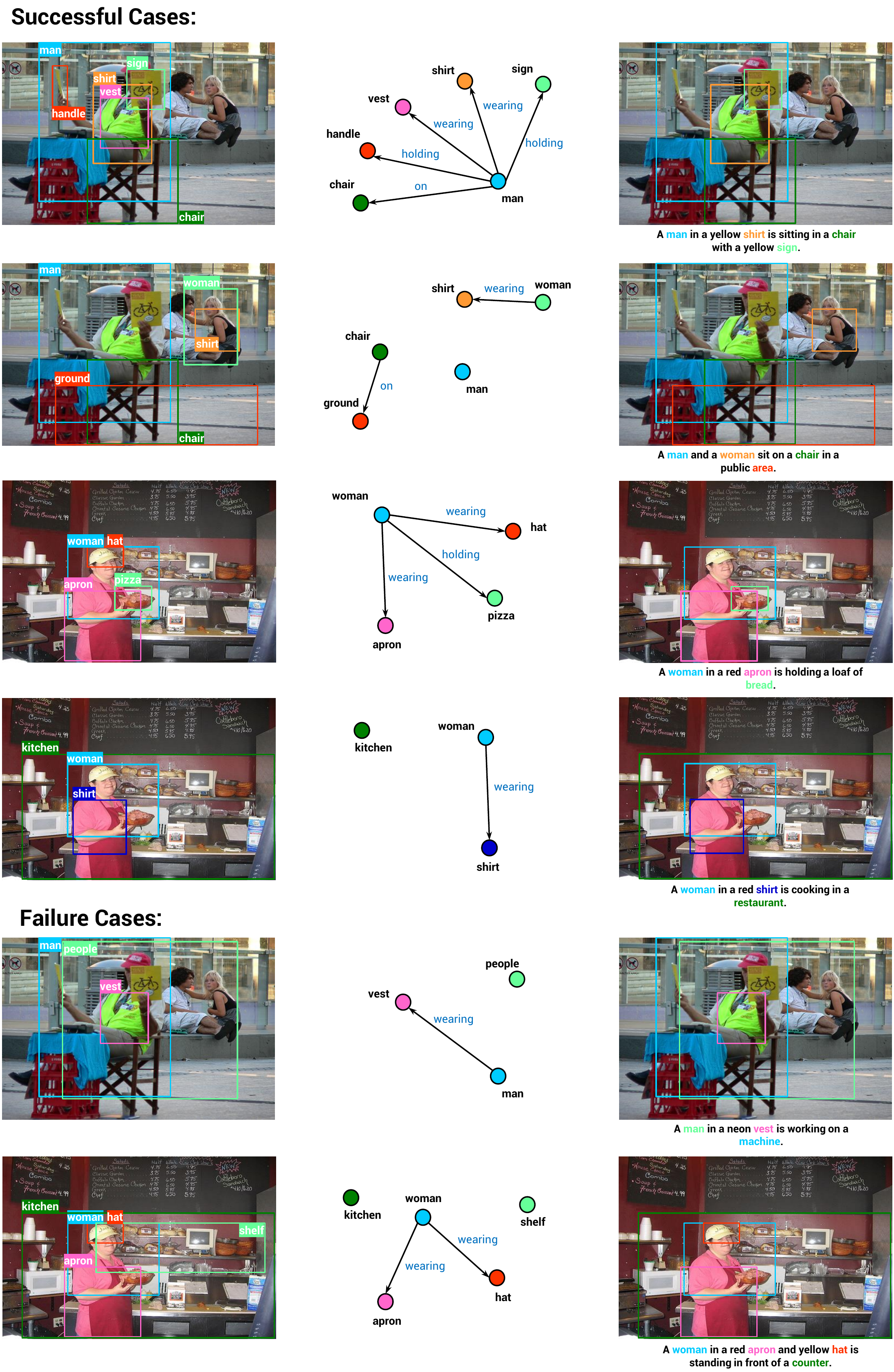}
    \caption{Diverse and grounded captioning results on Flickr30k Entities test set (Part 4). Each row presents the result from a single sub-graph. From left to right: input image and detected objects associated with the sub-graph, sub-graph nodes and edges used to decode the sentences, and the generated sentence grounded into image regions. Decoded nouns and their corresponding grounding regions are shown in the same color.}\label{fig:subgraph_grounding_visualization_part4}
\end{figure*}

We present further qualitative results of our model on Flickr30k Entities test set. Given an input image and its scene graph, our method selects multiple top ranked sub-graphs, decodes each of them into a sentence description and associates the decoded sentence tokens with the image regions. Sample results of the selected sub-graphs, decoded sentences and their region groundings are visualized in \Cref{fig:subgraph_grounding_visualization_part1,fig:subgraph_grounding_visualization_part2,fig:subgraph_grounding_visualization_part3,fig:subgraph_grounding_visualization_part4}. 

For each image, we show multiple generated sentences decoded from different sub-graphs grouped into successful and failure cases. For each row, we present results from a single sub-graph, including its detected objects (left), the nodes and edges used to decode the sentence (middle), and the output sentence grounded into image regions (right). For each sub-graph, we only visualize the nodes that have maximum attention weights for the decoded tokens, as well as the edges between these nodes (middle). Moreover, we present the decoded nouns and their grounded image regions using the same color (right). 

Take the first image shown in Fig.\ \ref{fig:subgraph_grounding_visualization_part1} as an example. Our model describes this image as ``A man in a suit is walking down the street'' when using a sub-graph with the nodes of ``man'', ``jacket'' and ``sidewalk'', or as ``A bus is parked in front of a building'' when using another sub-graph with nodes of ``bus'' and ``building''. Moreover, our model successfully links the generated tokens, such as ``man'', ``suit'', ``street'', ``bus'' and ``building'' to their image regions. These results further suggest that our method can generate diverse and grounded captions by representing scene components as sub-graphs on an input scene graph. 

\end{appendices}

\end{document}